\RequirePackage{amsmath}
\documentclass[a4paper]{llncs}

\usepackage[american]{babel}

\usepackage{graphicx}
\usepackage{amssymb}
\usepackage{amsmath}
\usepackage{url}
\DeclareFontFamily{U}{MnSymbolC}{}
\DeclareSymbolFont{MnSyC}{U}{MnSymbolC}{m}{n}
\DeclareFontShape{U}{MnSymbolC}{m}{n}{
    <-6>  MnSymbolC5
   <6-7>  MnSymbolC6
   <7-8>  MnSymbolC7
   <8-9>  MnSymbolC8
   <9-10> MnSymbolC9
  <10-12> MnSymbolC10
  <12->   MnSymbolC12%
}{}
\DeclareMathSymbol{\powerset}{\mathord}{MnSyC}{180}

\usepackage{color}

\usepackage{booktabs}

\begin{document}

\input glyphtounicode.tex
\pdfgentounicode=1

\title{(Blue) Taxi Destination and Trip Time Prediction from Partial Trajectories}

\author{Hoang Thanh Lam, Ernesto Diaz-Aviles,\\ Alessandra Pascale, Yiannis Gkoufas, and  Bei Chen}
\authorrunning{Hoang Thanh Lam et al.}
\institute{IBM Research -- Ireland\\
\vspace{1em}
\email{\{t.l.hoang, e.diaz-aviles, apascale, yiannisg, beichen2\}@ie.ibm.com}
}
%
%
			
\maketitle

\begin{abstract}
Real-time estimation of destination and travel time for taxis is of great importance for existing electronic dispatch systems. We present an approach based on trip matching and ensemble learning, in which we leverage the patterns observed in a dataset of roughly 1.7~million taxi journeys to predict the corresponding final destination and travel time for ongoing taxi trips, as a solution for the ECML/PKDD Discovery Challenge 2015 competition. The results of our empirical evaluation show that our approach is effective and very robust, which led our team --~BlueTaxi~-- to the 3rd and 7th position of the final rankings for the trip time and destination prediction tasks, respectively. Given the fact that the final rankings were computed using a very small test set (with only 320 trips) we believe that our approach is one of the most robust solutions for the challenge based on the consistency of our good results across the test sets.
\end{abstract}


\section{Introduction}
Taxi dispatch planning and coordination is challenging given the traffic dynamics in modern urban cities. The improvement of electronic dispatching of taxis can have a positive impact in a city, not only by reducing traffic congestion and environmental impact, but also improving the local economy, e.g., by mobilizing more customers in cities where the demand of taxis is very high. 

Current electronic dispatch systems capture localization and taximeter state from mobile data terminals installed in GPS-enabled taxis. The collective traffic dynamics extracted from the taxis' GPS trajectories and the stream of data collected in real-time represent a valuable source of information to understand and predict the behavior of future trips. In particular, we are interested in improving taxi electronic dispatch by predicting the final destination and the total trip time of ongoing taxi rides based on their initial partial trajectories. A solution to this challenge is very valuable given that in most cases busy taxi drivers do not report the final destination of the ride, which makes electronic dispatch planning more difficult and sub-optimal, e.g., ideally the destination of a ride should be very close to the start of the next one.

These two tasks, namely taxi destination and trip time prediction, are the ones that correspond to the Discovery Challenge part of ECML/PKDD 2015~\cite{ecmlpkdddc}. This paper details our data-driven solution to the challenge that placed our team, \emph{BlueTaxi}, in the 3rd and 7th position in the final ranking for the time and destination prediction task, respectively.

In particular, we are given a data set with roughly 1.7 million taxi journeys from 442 taxis operating in the city of Porto, Portugal, for a  period of one year. Besides polylines with GPS updates (at a resolution of 15 seconds), for every taxi trip we are provided with additional information that includes the taxi id, origin call id, taxi stand id, starting timestamp of the ride, etc. The competition is hosted on Kaggle's platform\footnote{\url{https://www.kaggle.com/}}, which allows participants to submit predictions for a given test set whose ground-truth is held by the competition organizers. Submissions are scored immediately (based on their predictive performance relative to a 50\% of the hidden test dataset) and summarized on a live \emph{public} leaderboard (LB). At the end of the competition, the other 50\% of the held-out test set, i.e., from which the participants did not receive any feedback, is used to compute the final rankings, also known as the \emph{private} LB. The challenge details and dataset characteristics can be found in~\cite{ecmlpkdddc}.

The challenge organizers explicitly asked for prediction of trip destination and duration at 5 specific time snapshots, namely 2014-08-14 18:00:00, \mbox{2014-09-30} 08:30:00, 2014-10-06 17:45:00, 2014-11-01 04:00:00, and 2014-12-21 14:30:00.

\begin{figure}[bt]
      \centering
      \includegraphics[width=0.85\textwidth]{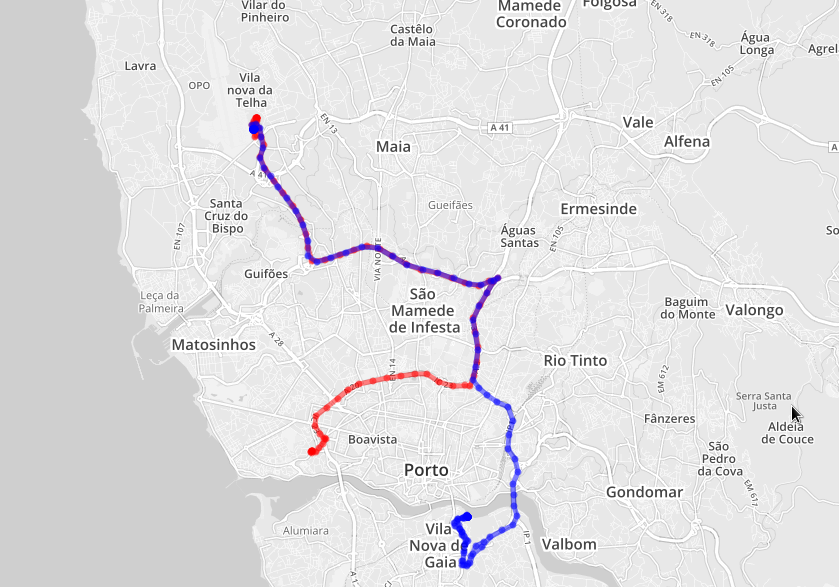}
      \caption{Two trips with different starting points but with the same destination (Porto airport). The final part of the trajectories are very close to each other, which helps to estimate the destination of similar trips.}
      \label{knn}
  \end{figure}

Our solution is based on a simple intuition: two trips with similar route likely end at the same or nearby destinations. For example, Figure \ref{knn} shows two complete trips extracted from the data. Both trips ended at the Porto airport. Although the first parts of the trips are very different, the last parts of the trips are the same. Therefore, using the destinations of similar trips in the past we can predict the destination of a test trip. 

This simple intuition serves as the key idea behind our approach in which we extract features (destination and trip time) from similar trips and then build a model ensemble based on the constructed features to predict taxis' destination and travel time. In the next sections we detail the feature extraction process, the predictive models of our approach, and present the experimental results that show the effectiveness of our solution.

\section{Feature Extraction}
\label{sec:feature_extraction}

From our initial data exploration, we found that the dataset contains missing GPS updates and erroneous information. In fact, the \emph{missing value} column part of the dataset was erroneous: most values of the column were \emph{False}, even though the records included missing information. We detected missing GPS points in the trajectory by observing large jumps between two consecutive GPS updates, i.e., by considering distances exceeding the distance that a taxi would have travelled at a speed limit (e.g., 160~km/h). 

We also found that the trip start timestamp information was unreliable, e.g., some trips in the test set have starting timestamps 5 hours before the cut-off timestamps of the corresponding snapshot but they contain only a few GPS updates. Moreover, some taxi trips have very unusual trajectories, e.g., we noted that some taxi drivers forgot to turn the meter off after completing a trip, for instance in the way back to the city center after dropping a passenger at the airport, such \emph{turning back} trajectories were very difficult to predict.

Considering these issues, we preprocessed the data accordingly and extracted the features that are detailed in this section, which will serve as input to build our predictive models for destination and trip time prediction.
  
\subsection{Feature Extraction for Destination Prediction}
For predicting the destination of a given incomplete test trip $A$, our method is based on trip matching which uses destinations of trips in the training data with similar trajectories to predict the final destination of the given test trip. Figure~\ref{knn} shows an example of two trips with different starting points both going toward the Porto airport. As we can see, the final part of the trips are very similar, both trips took the same highways before ending at the same destination. We captures such pattern by a set of features described as follows.

\noindent -- Final destination coordinates and Haversine distance to 10 nearest neighbors: for a given test trip $A$ we search for 10 trips in the training data that are closest to it. Similarly, for each pair of trips $A$ and $B$ we compute the mean Haversine distance (Equation~\ref{eq:haversine}) between the corresponding points in their trajectories. We ignored trips $B$ that have fewer GPS updates than the ones in $A$.

\noindent -- Kernel regression (KR) as a smooth version of k-NN regression method. Our previous work on bus arrival time prediction shows that KR gives better results than k-NN~\cite{ecmlpkdd2015,sinn2012predicting}. Destinations calculated by KR were used as features to estimate the final destination of a test trip. KR requires to set the bandwidth parameter, in our experiment, we set it to 0.005, 0.05 and 0.5 to obtain three different \mbox{KR-based} destination predictions corresponding to these values.

\noindent -- The aforementioned features from KR are sensitive to the metric used for evaluating the similarity between trips. We experimented with dynamic time warping and mean Haversine distance, but only the latter metric was chosen given its efficiency in computation. Moreover, some trips may have very different initial trajectory but still share the same destination, e.g., see Figure \ref{knn}. Therefore, besides computing KR predictions for the full trip, we also compute them using only the last $d$ meters of the ride during the trip matching step, where $d \in $ \{100, 200, 300, 400, 500, 700, 1000, 1200, 1500\}. The models showed that the last 500 to 700 meters of the trips are the most important features for predicting both latitude and longitude of the final destination.

\noindent -- For a pair of trips $A$ and $B$, we also consider the distance metric that looks for the best match along their trajectories without alignment. The empirical results showed that KR predictions by best matching gives more accurate results than first aligning  the initial part of $A$'s trajectory with $B$'s.

\begin{figure}[bt]
      \centering
      \includegraphics[width=0.85\textwidth]{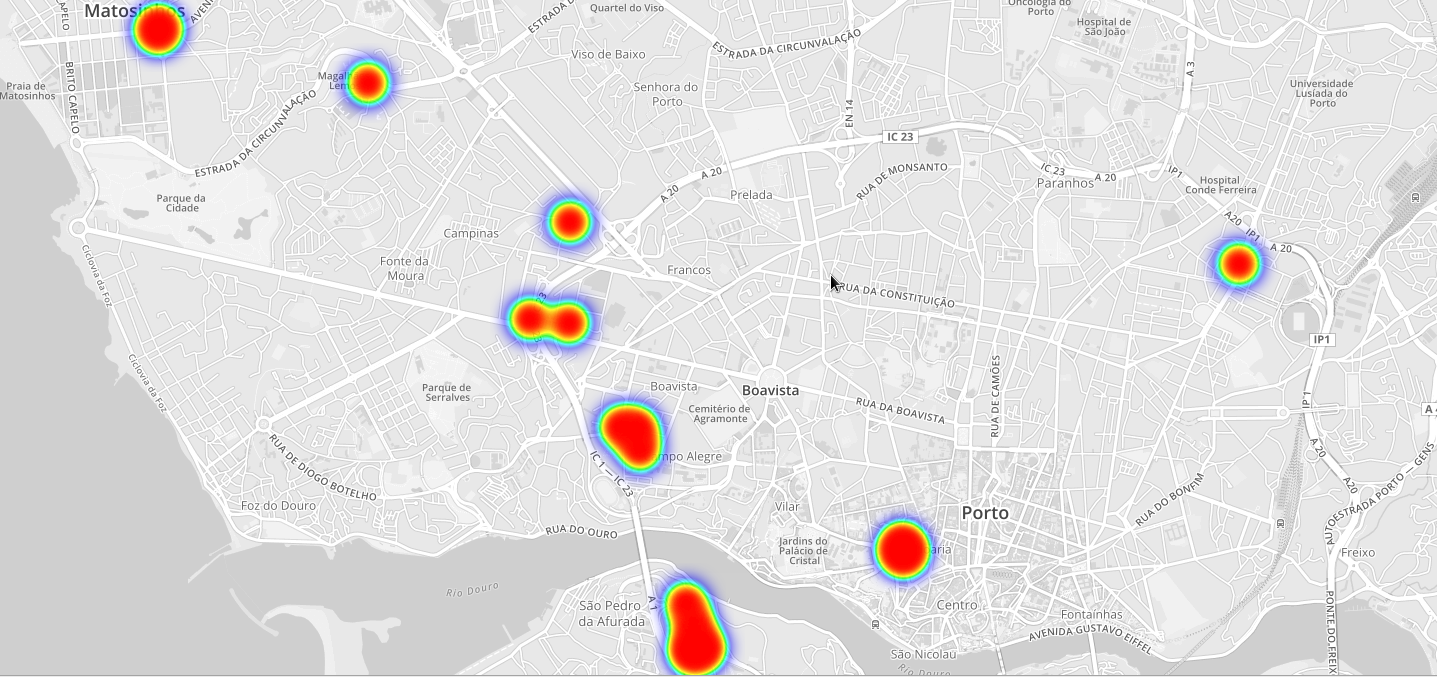}
      \caption{Heatmap of destinations of 155 trips with call id 59708. There are only 8 major destinations. Using call id information one can narrow down the possible destinations of a given taxi.}
      \label{fig:call_id}
\end{figure}

\noindent -- The features described so far do not consider contextual information such as the taxi id, call id, taxi stand, time of day, or day of the week, which are very important, consider for example Figure~\ref{fig:call_id}, which shows a heatmap of the destinations for all trips with origin call id 59708 in the training data, as we can see the destination is quite regular, although overall, there are 155 trips, only 8 major destinations were identified. In our work, we exploit this information by using KR to match a test trip with only trips with the same call id, taxi id, day of the week, hour of the day, or taxi stand id. We called the KR models based on these features \emph{contextual KR}. When a contextual field of the test trip has a missing value, the corresponding contextual KR was replaced by the KR prediction with no contextual information. The empirical results showed that contextual \mbox{KR-based} prediction on the call id gave the best results among all KR contextual related features.

\noindent -- We observed that there are many GPS updates that are erroneous, e.g., coordinates completely outside the trajectory shape or some GPS updates are very close together due to traffic jam or when a taxi is parked or stays still at the same location with the meter on. These type of noisy GPS updates influence the performance of the KR prediction, especially the KR using the mean Haversine distance metrics which requires point to point comparison across trips. Therefore we used the RDP algorithm~\cite{douglas:73,Ramer1972244} to simplify the trips with parameter $\epsilon$ equal to $1\times10^{-6}$, $5\times10^{-6}$, and $5\times10^{-5}$. From the simplified trips we extracted all KR-related features as we did for the \emph{raw} trip trajectories. 

Besides features extracted via trip matching we also added features extracted directly from the partially observed trips:

\noindent -- Euclidean distance traveled and Haversine distance between the first and the last GPS update.
  
\noindent -- Direction: to define whether the taxi is moving outside of the city or vice versa. We compared the distance between the city center and the first and the last point of the trips. If the former is larger the taxi is considered as moving toward the city center and moving away (e.g., to the countryside) otherwise.

\noindent -- Time gap between the cut-off timestamps and the starting timestamps. We observed that this feature is not very reliable because the starting timestamps are quite noisy.  

\noindent -- Number of GPS updates (this feature is also noisy due to missing values).

\noindent -- Day of the week.  We observed that the prediction error was higher for some days of the week, but the Random Forest model used in our experiments did not rank this feature high for destination prediction (cf. Section~\ref{sec:results}). 

\noindent -- The first and the last GPS location.
   
\subsection{Feature Extraction for Trip Time Prediction}
\label{sec:feature_extraction_for_trip_prediction}
The set of features for the time prediction task is very similar to the set of features for destination prediction, with the difference that travel time of closest trips were considered as the target variable instead of destination. The features extracted for this task are as follows.

\noindent -- Travel time and Haversine distance to 10 nearest neighbors.

\noindent -- Kernel regression features. All KR related features for destination prediction were also extracted for time prediction, with the difference that \emph{travel time} of closest trips was considered as target variable instead of destination.

In addition to, all features extracted from the partially observed trips as described in the previous section, we also considered the following additional particular features for time prediction which were extracted directly from the incomplete trips observed (i.e., trips that are still ongoing):

\noindent -- Average speed calculated on the last $d$ meters of the partial trajectory observed so far and on the entire incomplete trip, where $d \in $ \{10, 20, 50, 100, 200\}. These features convey up-to-date traffic condition at the moment of making a prediction.

\noindent -- Average speed calculated on all the incomplete trips observed so far with the starting time no more than an hour apart from the cut-off time-stamp. These features convey information about the traffic condition around the snapshot timestamps.

\noindent -- Average acceleration calculated  on the last $d$ meters of the incomplete trips observed so far and on the entire incomplete trips, where $d \in $ \{10, 20, 50, 100, 200\}.

\noindent -- Shape complexity: the ratio between the (Euclidean) traveled distance and the Haversine distance between the first and the last GPS location. Trips with higher complexity (e.g., zig-zag trips) tend to be trips for which the taxi drivers were driving around the city to search for passengers. Zig-zag trips tend to have longer travel time so it is reasonable to identify those trips beforehand.

\noindent -- Missing values in the GPS trace were identified by calculating the speed between any pair of consecutive GPS updates. If the estimated speed is over the speed limit $\hat{v}$ km/h even for only one pair of consecutive GPS updates in the partially observed trip, the trip was labelled as a trip with missing GPS updates. We used speed limits $\hat{v} \in $ \{100, 120, 140, 160\} km/h. Trips with missing values tend to have longer travel time.  

In total, we have 66 features for the trip time prediction tasks.

\section{Predictive Modelling}
In order to train the predictive models, first, we created a local training dataset by considering 5 time snapshots from weekdays that resemble the disclosed test set and was fixed for all trained models.

Overall we extracted 13301 trips from these snapshots in the training data, which is a small fraction of the 1.7 million trips in the original training set. We also considered roughly 12000 additional trips from 5 snapshots with start timestamps one hour after the specified test set snapshots. However, models trained on these combined sets yielded improvement only locally but not on the public leaderboard. Therefore, we decided to ignore the additional training trips although in practice to obtain robust results one should consider these additional data for training the models as well. 
       
We observed that feature extraction for the training set is not very efficient because most features were extracted based on nearest neighbor search. In order to speed up this process, we propose to use an index structure based on \emph{geohash}\footnote{\url{https://en.wikipedia.org/wiki/Geohash}}. We first represent each GPS by its geohash, then we search for the nearest trips via range queries to retrieve trips with a maximum distance threshold of 1km from the first point of the test trip. This simple indexing technique speeds up the nearest neighbor search significantly since range queries are very efficient with geohashes. It is important to note that this technique is a heuristic solution which does not guarantee exact nearest neighbors results. However, we did not observe significant differences in the prediction results when exact or approximated nearest neighbors were used.

In the rest of this section, we describe in detail the different models for each of the prediction tasks.

\subsection{Models for Destination Prediction}
Given the set of features, any regression model can be used to predict latitude and longitude independently. To this end, we chose Random Forest (RF)~\cite{rf} given the robustness of its predictions observed in both the local validation set and in the public LB test set. Besides RF models, we also experimented with Support Vector Regression (SVR), but the results on the public LB did not differ much, so we used only RF for the final destination prediction. Moreover, with a RF model we can easily assess the contribution of each feature on the final prediction. 
With this insight, we know whether a new set of features is relevant or not every time it is added to the model. Thanks to this we could perform feature selection using the \texttt{rfcv} function provided with the \texttt{randomForest} package in R. With feature selection the obtained results did not differ on the public LB but the results were more robust, in fact it needs fewer trees to obtain similar prediction results.  
  
To handle outliers, we first trained an initial RF model with 2000 trees and removed from training set all trips that prediction error was greater than 90\% of error quantile. Subsequently, a new RF model  re-trained  with the new training set became our final model.

\subsection{Ensemble for Trip Time Prediction}
For the trip time prediction, before training the models we removed outliers with a trip travel time that exceeded a median absolute deviation of 3.5. Furthermore, we trained the models using as target variable the (log-transformed) \emph{delta time} between the cut-off time point of our training snapshot and the timestamp associated to the last point of the trajectory. That is, our goal is to train models to predict the trip time remaining for a given ongoing trip. This preprocessing strategy led to significantly better results.

Then, we used the features described in Section~\ref{sec:feature_extraction_for_trip_prediction} to train a model ensemble for a robust prediction. The individual members of the ensemble include the following regression models: Gradient Boosted Regression Trees~(GBRT)~\cite{gbrt}, Random Forest regressor~(RF)~\cite{rf}, and Extremely Randomized Trees regressor~(ERT)~\cite{ert}.

In order to produce a single predictor we follow a \emph{stacked generalization (stacking)}~\cite{wolpert199} approach summarized as follows:

\noindent -- Remove from our training data, $D_{T}$, a subset of the samples (i.e., trips) and split them equally to form a validation set $D_{v}$  and (local) test set $D_{t}$, i.e., $|D_{v}| = |D_{t}|$.

\noindent -- Train $n$ models on the remaining training data $D^{'}_{T} = D_{T} \setminus (D_{v} \cup D_{t})$.

\noindent -- For each model, compute the predictions for the validation set $\mathbf{P}^v$.

\noindent -- For the validation set the corresponding true trip time is known, which leads to a supervised learning problem: using the predictions $\mathbf{P}^v_n$ as the input and the correct responses in $D_{v}$ as the output, train a meta-regressor $M_e(\mathbf{P}^v_n)$ to ensemble the predictions.

\noindent -- Insert the validation set $D_{v}$ back into the training data $D^{'}_{T}$.

\noindent -- Train the models again using the $D^{'}_{T} \cup D_{v}$ training dataset.

\noindent -- Predict for the test set $D_{t}$ to obtain $\mathbf{P}^t$.

\noindent -- Ensemble the final prediction for the test set using $M_e(\mathbf{P}^t_n)$.

Table~\ref{tab:models_trip_time_prediction} details the models parametrization and predictive performance in terms of the Root Mean Squared Logarithmic Error (RMSLE) defined in Equation~\ref{eq:rmsle}. We experimented with regularized linear regression and with simple average for the meta-regressor to produce the final prediction. The results of our approach are presented in Section~\ref{sec:results_trip_time_prediction}.

\section{Experimental Results}
\label{sec:results}
During the competition, the performance on the ground-truth test set held by the organizers was only available for a 50\% of the test trips and via a limited number of submissions to the Kaggle's website. Therefore, in order to analyze the predictions results locally, we split our subset of training data with 13301 trips into two parts: local training and validation datasets. 
This section reports the experimental results for the public and private LB test sets as well as for the local validation set.

\subsection{Destination Prediction}
The evaluation metric for destination prediction is Mean Haversine Distance (MHD), which measures distances between two points on a sphere based on their latitude and longitude. The MHD can be computed as follows.
\vspace{-0.25em}
\begin{align}
\textbf{MHD} &= 2 \cdot r \cdot \arctan\big( \sqrt{\frac{a}{1 - a}} \big) \text{ , where}
\label{eq:haversine}\\
a  &= \sin^2\big( \frac{\phi_2 - \phi_1}{2} \big) + \cos(\phi_1)\cdot\cos(\phi_2)\cdot\sin^2\big( \frac{\lambda_2 - \lambda_1}{2} \big) \notag
\end{align}
and $\phi$, $\lambda$, $r=6371$km are the latitude, longitude, and the Earth's radius, respectively.

\begin{table}[!t]
\caption{Destination prediction error in terms of the Mean Haversine Distance (MHD).}
\label{tab:mhd}
\centering
\setlength{\tabcolsep}{1em}
\scalebox{0.8}{
\begin{tabular}{l c}
\multicolumn{2}{c}{(a)} \\
\toprule
\textbf{Dataset}    & \textbf{MHD (km)}\\
\midrule
Public LB  & 2.27 \\
Private LB & 2.14 \\
Local validation set & 2.39\\
&\\
&\\
\bottomrule
\end{tabular}
\quad
\begin{tabular}{l c}
\multicolumn{2}{c}{(b)} \\
\toprule
\textbf{Day of the week and time}    & \textbf{MHD (km)}\\
\midrule
Monday   \;\;\,(17:45)    & 2.24 \\
Tuesday  \;\;\,(08:45)    & 2.40 \\
Thursday \,(18:00)  & 2.48 \\
Saturday \;\,(04:00)  & 2.21 \\
Sunday   \quad(14:30)    & 2.95 \\
\bottomrule
\end{tabular}
}
\vspace{-0.5em}
\end{table}

\begin{figure}[!t]
    \centering
    \includegraphics[width=0.95\textwidth]{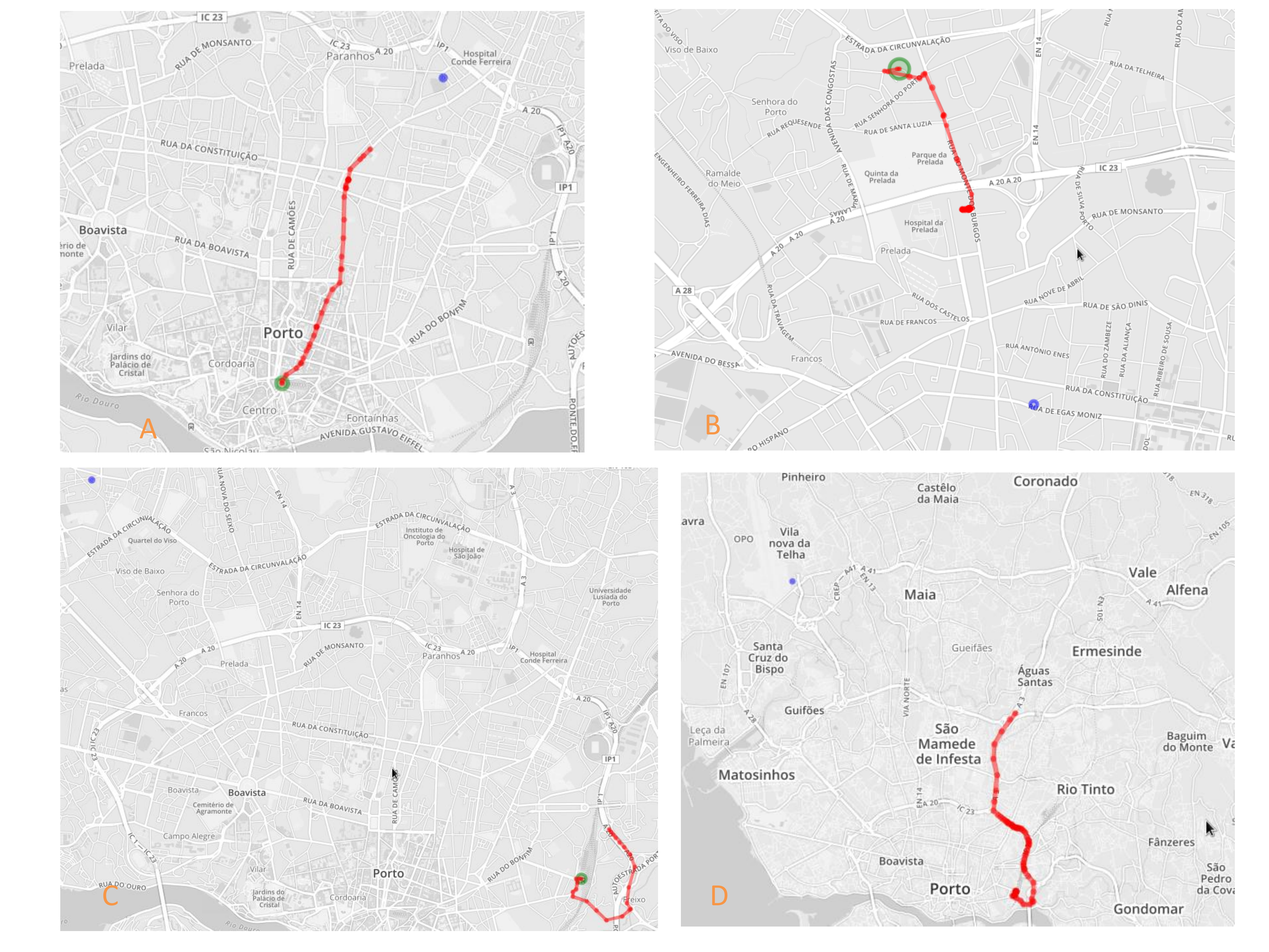}
    \caption{Examples of trips on the LB test set and our prediction (blue circle).}
    \label{prediction}
\end{figure}

Table~\ref{tab:mhd}(a) shows the mean Haversine distance from the predicted destination to the correct destination on three different test sets. Note that for this task we used a \mbox{66\%--34\%} training--validation split. We can see that the results are quite consistent and the prediction errors on three different test sets are small. 

Table~\ref{tab:mhd}(b) reports the prediction error on the local test set at different snapshots. As we can see, among the snapshots, trips in early morning on Saturday are easier to predict while trips on Sunday afternoon are very difficult to predict with significantly higher prediction error. On Sundays, taxis tend to commit longer trips with irregular destinations while early morning Saturdays taxis usually go to popular destinations like airports, train stations, or hospitals.

Figure~\ref{prediction} shows four trips on the LB test set and our prediction. Some predictions are just about one block away from the last GPS location (cases A and B), while some predictions are very far away from the last location (cases C and D). The later cases usually concern trips that took the highway to go to the other side of the city by skipping the city center or go to the airport via A3 motorway.

\subsection{Trip Time Prediction}
\label{sec:results_trip_time_prediction}
For the trip time task, predictions are evaluated using the Root Mean Squared Logarithmic Error (RMSLE), which is defined as follows:
\begin{equation}
\textbf{RMSLE} = \sqrt{\frac{1}{n} \sum_{i=1}^{n}{(\ln(p_i + 1) - \ln(a_i + 1))^2}}
\label{eq:rmsle}
\end{equation}
where $n$ is the total number of observations in the test data set, $p_i$ is the predicted value, $a_i$ is the actual value for travel time for trip $i$, and $\ln$ is the natural logarithm.

Table~\ref{tab:models_trip_time_prediction} summarizes the predictive performance of the individual model members of our ensemble locally using a 80\%--20\% training--validation split.

\begin{table}[!htb]
\caption{Regression models for trip time prediction and the corresponding performance (RMSLE) in our local test dataset. The models include Gradient Boosted Regression Trees~(GBRT), Random Forest regressor~(RF), and Extremely Randomized Trees regressor~(ERT).}
\label{tab:models_trip_time_prediction}
\centering
\setlength{\tabcolsep}{0.5em}
\scalebox{0.6}{
\begin{tabular}{l p{40em} r}
\toprule
\textbf{Model} & \textbf{Parameters} & \textbf{RMSLE}\\
\toprule
\toprule
\textbf{GBRT} & Default parameter settings used for the GBRT models: learning\_rate=0.1, max\_depth=3, max\_features=n\_features, min\_samples\_leaf=3, min\_samples\_split=3, n\_estimators=128 & \\
\midrule
GBRT-01 & loss=`squared\_loss', subsample=1.0 & 0.41508\\
\midrule
GBRT-02  & loss=`squared\_loss', subsample=0.8 & 0.41498\\
\midrule
GBRT-03  & loss=`least\_absolute\_deviation', subsample=1.0 & 0.40886\\
\midrule
GBRT-04 & loss=`least\_absolute\_deviation', subsample=0.8 & 0.40952\\
\midrule
GBRT-05  & loss=`huber', subsample=1.0, alpha\_quantile=0.9& 0.41218\\
\midrule
GBRT-06 & loss=`huber', subsample=0.8, alpha\_quantile=0.9& 0.41108\\
\midrule
GBRT-07 & loss=`huber', subsample=1.0, alpha\_quantile=0.5& 0.41000\\
\midrule
GBRT-08 & loss=`huber', subsample=0.8, alpha\_quantile=0.5& 0.40705\\
\midrule
GBRT-09  & loss=`quantile', subsample=1.0, alpha\_quantile=0.5& 0.40798\\
\midrule
GBRT-10 & loss=`quantile', subsample=0.8, alpha\_quantile=0.5& 0.40959\\
\toprule
\toprule
\textbf{RF} & Default parameter settings used for the RF models: max\_depth=None, n\_estimators=2500, use\_out\_of\_bag\_samples=False& \\
\midrule
RF-01 & max\_features=n\_features, min\_samples\_leaf=4, min\_samples\_split=2  & 0.41674\\
\midrule
RF-02 & max\_features=n\_features, min\_samples\_leaf=1, min\_samples\_split=1& 0.41872\\
\midrule
RF-03 & max\_features=sqrt(n\_features),min\_samples\_leaf=4, min\_samples\_split=2& 0.41737\\
\midrule
RF-04 & max\_features=$\log_2$(n\_features),min\_samples\_leaf=4, min\_samples\_split=2& 0.41777\\
\midrule
RF-05 & max\_features=n\_features,min\_samples\_leaf=1,min\_samples\_split=1, use\_out\_of\_bag\_samples=True & 0.41865\\
\midrule
RF-06 & max\_features=sqrt(n\_features),min\_samples\_leaf=4,min\_samples\_split=2, use\_out\_of\_bag\_samples=True & 0.41731\\
\midrule
RF-07 & max\_features=$\log_2$(n\_features),min\_samples\_leaf=4,min\_samples\_split=2, use\_out\_of\_bag\_samples=True & 0.41790\\
\toprule
\toprule
\textbf{ERT} & Default parameter settings used for the RF models: max\_depth=None, n\_estimators=1000, use\_out\_of\_bag\_samples=False& \\
\midrule
ERT-01 & max\_features=n\_features, min\_samples\_leaf=1, min\_samples\_split=1 & 0.41676\\
\midrule
ERT-02 & max\_features=n\_features, min\_samples\_leaf=1, min\_samples\_split=2 & 0.41726\\
\midrule
ERT-03 & max\_features=n\_features,min\_samples\_leaf=1,min\_samples\_split=2, use\_out\_of\_bag\_samples=True & 0.41708\\
\midrule
ERT-04 & max\_features=n\_features,min\_samples\_leaf=1,min\_samples\_split=1, use\_out\_of\_bag\_samples=True, n\_estimators=3000 & 0.41735\\
\bottomrule
\end{tabular}
 }
\end{table}
\begin{table}[!tb]
\caption{Trip time prediction results on the public and private leaderboards in terms of RMSLE, where the lower the value, the better.}
\label{tab:trip_time_prediction_results}
\centering
\setlength{\tabcolsep}{1em}
\scalebox{0.7}{
\begin{tabular}{p{22em} r r r}
\toprule
\textbf{Model ensemble description} & \multicolumn{3}{c}{\textbf{RMSLE}}\\
 & \textbf{Public LB} & \textbf{Private LB} & \textbf{Average}\\
\midrule
ME1: Model ensemble using the average to compute the final prediction. & 0.49627 & 0.53097 & 0.51362\\
\midrule
ME2: Model ensemble using regularized linear regression with L2 penalty as meta-regressor to compute the final prediction. & 0.51925 & \textbf{0.51327} & 0.51626\\ 
\midrule
ME3: Model ensemble using regularized linear regression with L1 penalty (Lasso) as meta-regressor to obtain the final prediction. & 0.49855 & 0.52491 & \textbf{0.51173}\\
\midrule
ME4: Model ensemble trained using trips with instances whose number of GPS updates is between 2 and 612, which reassemble the distribution observed in the partial trajectories of the test set. The final prediction is computed using mean of the individual predictions. & 0.49418 & 0.54083 & 0.51750\\
\midrule
ME5: Average prediction of model ensembles ME1, ME2, M3, and M4  (\textbf{3rd place on the public LB}). & \textbf{0.49408}  & 0.53563 & 0.51485 \\
\midrule
ME6: Average prediction of model ensembles ME1, ME2 and M3  (\textbf{3rd place on the private LB}). & 0.49539 & 0.53097 &  0.51318\\
\bottomrule
\end{tabular}
}
\vspace{-2em}
\end{table}

In Table~\ref{tab:trip_time_prediction_results} we present our most relevant results in the official test set of the competition. The table shows the performance of our strategies in the public and private leaderboards, which corresponds to a 50\%--50\% split of the hidden test set available only to the organizers. Remember that the final ranking is solely based on the private leaderboard from which the participants did not receive any feedback during the competition.

We achieved the best performance in the public leaderboard (RMSLE=0.49408) using ME5, which corresponds to an average of model ensembles as specified in Table~\ref{tab:trip_time_prediction_results}, this result placed our team at the 3rd place. For the private leaderboard, we also achieved the top-3 final position using ME6 (RMSLE=0.53097). ME6 is similar to the ME5 approach, but it does not include in its average the predictions of ME4, which corresponds to the models trained using a subset of the trips with number of GPS updates in the range between 2 and 612 points.

Note that our ME2 strategy achieved a RMSLE of 0.51925 and 0.51327 in the public and private leaderboards, respectively. The RMSLE of this entry in the leaderboard would have achieved the top-1 position in the final rankings. The competion rules allowed to select two final sets of predictions for the final evaluation. Unfortunately, given the performance observed in the public leaderboard for which we had feedback, we did not select the ME2 approach as one of final two candidate submissions.

The model ensemble that uses regularized linear regression with L1 penalty (Lasso)~\cite{lasso} achieves the best average performance for the whole test dataset with an RMSLE=0.51173. 

We used \texttt{NumPy}\footnote{\url{http://www.numpy.org/}} and \texttt{scikit-learn}\footnote{\url{http://scikit-learn.org/}} to implement our approach for trip time prediction.
\section{Conclusions and Future work}
We propose a data-driven approach for taxi destination and trip time prediction based on trip matching. The experimental results show that our models exhibit good performance for both prediction tasks and that our approach is very robust when compared to competitors' solutions to the ECML/PKDD Discovery Challenge. For future work, we plan to generalize our current approach to automatically adapt to contexts and select a particular set of features that would improve predictive performance within the given context.

\bibliographystyle{splncs03}
\bibliography{2015_bluetaxi}

\end{document}